\documentclass{article}
\usepackage[ margin= 1 in]{geometry}
\usepackage{graphicx} 
\usepackage{caption} 
\usepackage{mathtools}
\usepackage{verbatim}
\usepackage{amsmath}
\usepackage{amsfonts}
\usepackage{float}
\usepackage{subfigure}
\usepackage{algorithm}
\usepackage[noend]{algpseudocode}
\usepackage{booktabs}
\usepackage{xparse, etoolbox, mathabx}
\usepackage[inline]{enumitem}
\usepackage[hyperfootnotes = false]{hyperref}

\usepackage[symbol]{footmisc}

\newcommand\blfootnote[1]{%
	\begingroup
	\renewcommand\thefootnote{}\footnote{#1}%
	\addtocounter{footnote}{-1}%
	\endgroup
}

\title{Semi-supervised sequence classification through change point detection}
\author{Nauman Ahad, Mark A. Davenport}

\begin{document}
\maketitle
\begin{abstract}
Sequential sensor data is generated in a wide variety of practical applications. A fundamental challenge involves learning effective classifiers for such sequential data. While deep learning has led to impressive performance gains in recent years in domains such as speech, this has relied on the availability of large datasets of sequences with high-quality labels. In many applications, however, the associated class labels are often extremely limited, with precise labelling/segmentation being too expensive to perform at a high volume. However, large amounts of unlabeled data may still be available.  In this paper we propose a novel framework for semi-supervised learning in such contexts. In an unsupervised manner, change point detection methods can be used to identify points within a sequence corresponding to likely class changes.  We show that change points provide examples of similar/dissimilar pairs of sequences which, when coupled with labeled, can be used in a semi-supervised classification setting. Leveraging the change points and labeled data, we form examples of similar/dissimilar sequences to train a neural network to learn improved representations for classification. We provide extensive synthetic simulations and show that the learned representations are superior to those learned through an autoencoder and obtain improved results on both simulated and real-world human activity recognition datasets.
\end{abstract}

\section{Introduction}

As devices ranging from smart watches to smart toasters are equipped with ever more sensors, machine learning problems involving sequential data are becoming increasingly ubiquitous. Sleep tracking, activity recognition and characterization, and machine health monitoring are just a few applications where machine learning can be applied to sequential data. In recent years, deep networks have been widely used for such tasks as these networks are able automatically learn suitable representations, helping them achieve state-of-the-art performance~\cite{wang2019deep}. However, such methods typically require large, accurately labeled training datasets in order to obtain these results. Unfortunately, especially in the context of sequential data, it is often the case that despite the availability of huge amounts of unlabeled data, labeled data is often scarce and expensive to obtain.

In such settings, \emph{semi-supervised} techniques can provide significant advantages over traditional supervised  techniques. Over the past  decade, there have been great advances in  semi-supervised learning methods. Impressive classification performance -- particularly in the fields of computer vision -- has been achieved by using large amounts of unlabeled data on top of limited labeled data. However, despite these advances, there has been comparatively much less work on semi-supervised classification of sequential data. 

A key intuition that most semi-supervised learning methods share is that the data should (in the right representation) exhibit some kind of clustering, where different classes correspond to different clusters. In the context of sequential data, the equivalent assumption is that data segments within a sequence corresponding to different classes should map to distinct clusters. In the context of sequential data, the challenge is that 
exploiting this clustering would require the sequence to be appropriately segmented, but segment boundaries are generally unknown \emph{a priori}. If the start/end points of each segment were actually known, it would be much easier to apply traditional semi-supervised learning methods.
\blfootnote{The authors are with the School of Electrical and Computer Engineering, Georgia Institute of Technology, Atlanta, GA, 30332 USA (emails: nahad3@gatech.edu, mdav@gatech.edu)}

In this paper, we show that standard (unsupervised) \emph{change point detection} algorithms provide a natural and useful approach to segmenting an unlabeled sequence so that it can be more easily exploited in a semi-supervised context. Specifically, change point-detection algorithms aim to identify instances in a sequence where the data distribution changes (indicating an underlying class change).
We show that the resulting change points can be leveraged to learn improved representations for semi-supervised learning.

We propose a novel framework for semi-supervised sequential classification using change point detection. We first apply unsupervised change point detection to the unlabeled data. We assume that segments between two change points belong to the same distribution and should be classified similarly, whereas adjacent segments which are on opposite sides of a change point belong to different distributions and should be classified differently. These similar/dissimilar pairs, derived from change points, can then be combined with  similar/dissimilar pairs derived from labeled data.  We use these combined similar/dissimilar constraints to train a neural network that preserves similarity/dissimilarity. 
The learned representation can then be fed into a multilayer feedforward network trained via existing semi-supervised techniques. 

We show that this approach leads to improved results compared to sequential auto-encoders 
in a semi-supervised setting.  
We show that even if the final classifier is trained using standard supervised techniques that ignore the unlabeled data, the learned representations (which utilize both label and unlabeled data pairs) result in competitive performance, indicating the value of incorporating change points to learning improved representations. The proposed method method is completely agnostic with respect to the  change point detection procedure to be used -- any detection procedure can be used as long as it does well in detecting changes.

Our main contribution is to show that pairwise information generated via change points helps neural networks achieve improved classification results in settings with limited labeled data. This, to the best of our knowledge, is the first work to recognize the utility of change points within the context of semi-supervised sequence classification.  The proposed method should not be considered a substitute for existing semi-supervised methods, but should be taken as a complementary procedure that produces representations which are better suited for existing semi-supervised methods.

\section{Related work}

The fundamental idea of semi-supervised learning is that unlabeled data contains useful information that can be leveraged to more efficiently learn from a small subset of labeled data.  For example, in the context of classification, an intuitive justification for why this might be possible might involve an implicit expectation that instances belonging to different classes will map to different clusters. More concretely, most semi-supervised approaches make assumptions on the data such as: that instances corresponding to different classes lie on different submanifolds, that class boundaries are smooth, or that class boundaries pass through regions of low data density~\cite{van2020survey}.

Perhaps the simplest semi-supervised learning method is to use transductive methods to learn a classifier on the unlabeled data and then assign ``pseudo labels'' to some or all of the unlabeled data, which can be used together with the labeled data to retrain the classifier. Transductive SVMs and graphical label propagation are examples of such methods~ \cite{joachims1999transductive,zhu2003semi}. See \cite{zhu2005semi} for a survey of such methods. However, such \emph{self-training} semi-supervised methods struggle when the initial model trained from limited labels is poor.  

A more common approach to semi-supervised learning 
is to employ methods that try to learn class boundaries that are smooth or pass  through areas of low  data density~\cite{oliver2018realistic}. Entropy regularization can be used to encourage class boundaries to pass through low density regions~\cite{grandvalet2005semi}. Consistency-based methods such as denoising autoencoders, ladder networks ~ \cite{rasmus2015semi} and the $\pi$ method~\cite{laine2016temporal} attempt to learn smooth class boundaries by augmenting the data. Specifically, unlabeled instances can be perturbed by adding noise, and while both the original and perturbed instances are unlabeled, we can ask that they both be assigned the same class.
This approach is particularly effective in computer vision tasks, where rather than using only noise perturbations, we can exploit class-preserving augmentations such as rotation, mirroring, and other transformations~\cite{berthelot2019mixmatch}. By enforcing the classifier to produce the same labels for original and transformed images, decision boundaries are encouraged to be smooth, leading to good generalization. 

Unfortunately, due to a lack of natural segmentation and the difficulty of defining class-preserving transformations, there has been comparatively little work on semi-supervised classification of sequences. Most prior work (e.g.,~\cite{dai2015semi, rasmus2015semi} ) use sequential autoencoders (or their variants) as a consistency-based method to learn representations that lead to improved classification performance. Such autoencoders have been exploited successfully in the context of semi-supervised classification for human activity recognition~\cite{zeng2017semi}. However, while such consistency-based approaches do encourage smooth class boundaries, they do not necessarily promote the kind of clustering behavior that we need in cases where there are extremely few labels available.

An alternative approach that more explicitly separates different classes involves learning representations that directly incorporate pairwise similarity information about different instances. One example of this approach is \emph{metric learning} -- as an early example, \cite{xing2003distance} showed that improved classification could be achieved by learning a Mahalanobis distance using pairwise constraints based on class membership. The learned metric leads to a representation in which different classes map to different clusters. A similar approach learns a more general non-linear metric to encourage the formation of clusters while adhering to the provided pairwise constraints \cite{baghshah2009semi}. Neural networks such as Siamese \cite{koch2015siamese} and Triplet networks also learn representations from available similar/dissimilar pairs.
In \cite{hsu2015neural} it was shown that such similar/dissimilar pairs (obtained from labeled data) can be used for clustering data  where each cluster belongs to a different class in the dataset.

Our approach is similar in spirit to that of \cite{hsu2015neural}. While this prior work used pairwise similarity constraints derived from labeled images to learn clustered representations, our goal is to apply this idea in the semi-supervised context. At the core of our approach is the observation that pairwise similarity constraints on sequential data can be derived through unsupervised methods. Specifically, change point detection can be used to identify points within a sequence corresponding to distribution shifts, which can then be used to obtain pairwise similarity constraints. When the availability of labeled data is limited, this can be a valuable source of additional information.


\begin{figure}
  \centering
  \includegraphics[width=8cm]{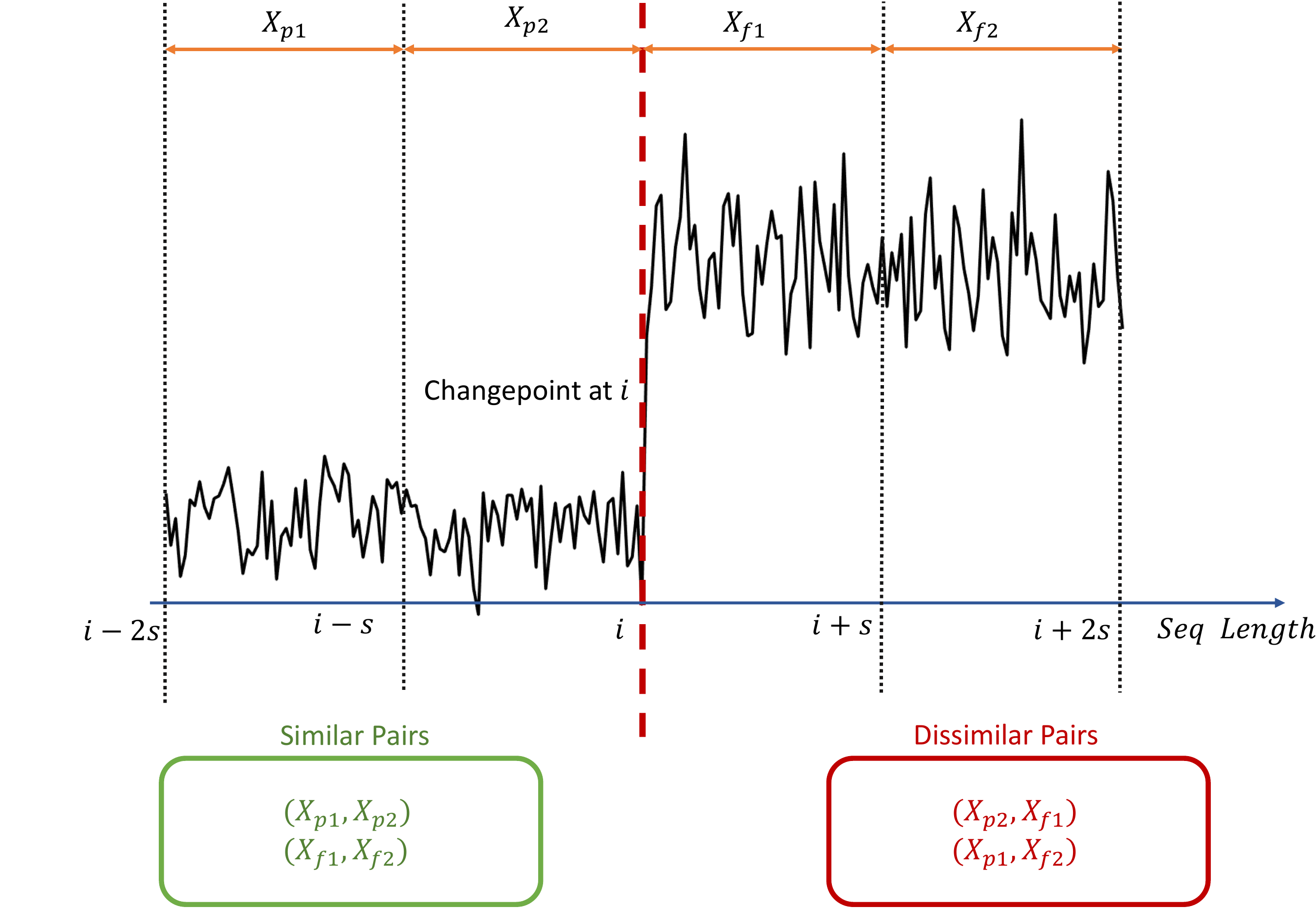}
  \caption{Using change points to generate similar and dissimilar pairs of size $s$.}
  \label{Fig: pairs from cp}
\end{figure}

\begin{figure*}[t]
  \centering
  \includegraphics[width=14cm]{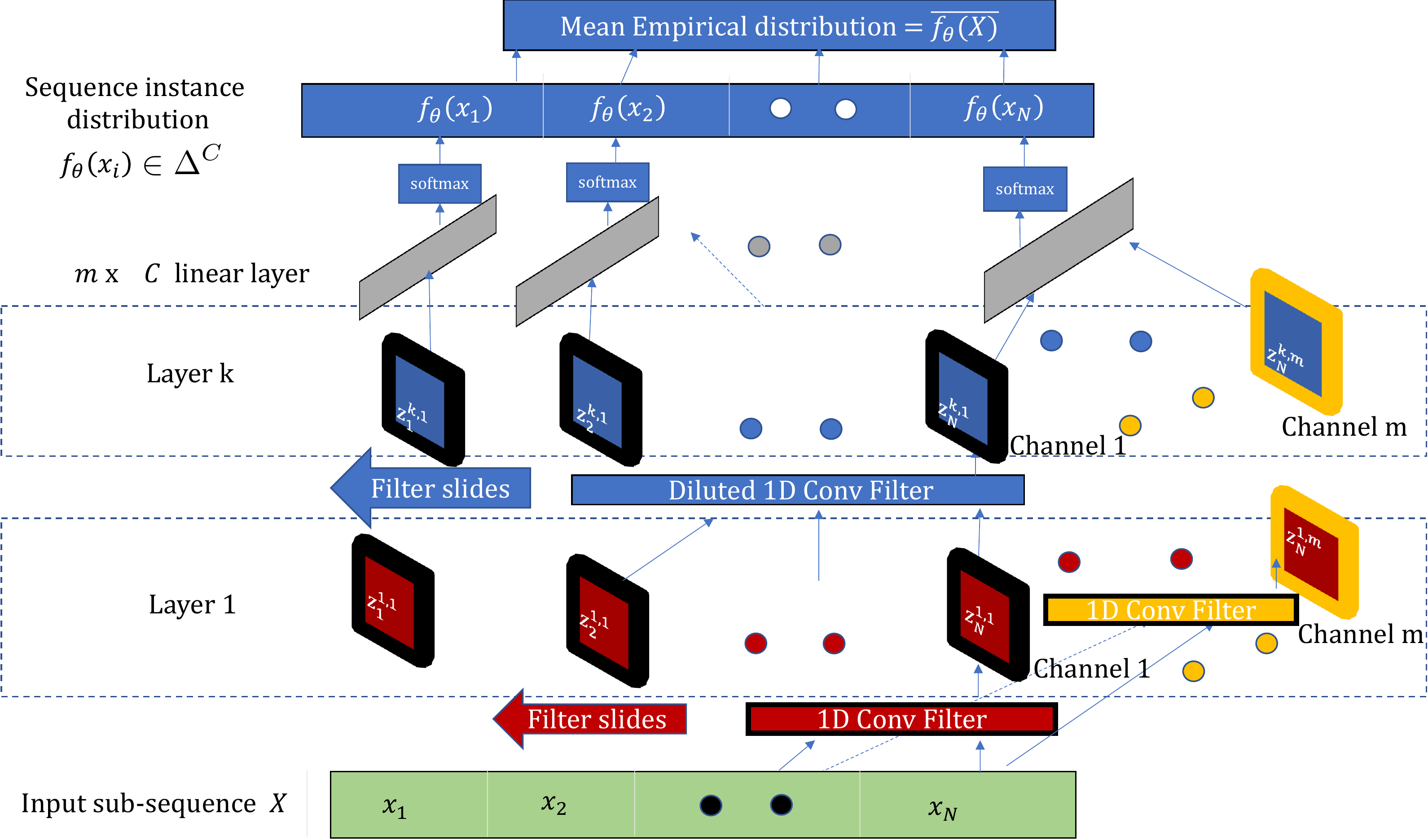}
  \caption{Neural network diagram ($f_{\theta}$) for learning representations.}
  \label{Fig: NN diagram}
\end{figure*}

\section{Proposed method}

\subsection{Change point detection}

Given a sequence $X: x_1, \ldots, x_N$ of $N$ vectors $x_i \in \mathbb{R}^D$, the first step in our procedure is to detect \emph{all} change points within $X$ in an unsupervised way. Note that this is a different problem than \emph{quickest} change detection, where only a \emph{single} change point is to be detected in the fastest possible manner.
To detect a change at a point $i$ in the sequence, two consecutive length-$w$ windows ($X_p^i$ and $X_f^i$)  are first formed:
\[
    X_p^i = x_{i-1}, x_{i-2}...x_{i-w} \quad \quad   X_f^i = x_{i}, x_{i+1}...x_{i+w}.
\]
A change statistic, $m_i$, is then computed via some function that quantifies the difference between the distributions generating $X_p^i$ and $X_f^i$.
If $m_i$ is greater than a specified constant $\tau$, a change point is detected at the point $i$.

As one example, many change point detection procedures assume a parametric form on the distributions generating $X_p^i$ and $X_f^i$. In this case, the distribution parameters ($\hat{\theta}_p^i$ and $\hat{\theta}_f^i$) can be estimated from $X_p^i$ and $X_f^i$ via, e.g., maximum likelihood estimation. Given these parameter estimates, 
a symmetrical KL-divergence can be used to quantify the difference between the distributions~\cite{liu2013change}: 
\begin{equation}
      m_i = \text{KL}( \hat{\theta}_p^i, \hat{\theta}_f^i ) +  \text{KL}( \hat{\theta}_f^i, \hat{\theta}_p^i ).
      \label{eq: Sym KL}
\end{equation}

More commonly in practice, the underlying distributions generating the sequence are unknown.  In this case, non-parametric techniques can be used to estimate the difference between the distributions of $X_p^i$ and $X_f^i$. One such approach uses the \emph{maximum mean discrepancy} (MMD) as a change statistic~\cite{gretton2012kernel}. The MMD has been used to identify change points in \cite{li2015m} and \cite{chang2018kernel}.  The MMD statistic is given below, where $K_{a,-b}^i : = k(x_{i+a},x_{i-b})$ represents a kernel-based measure of the similarity between $x_{i+a}$ and $x_{i-b}$:
\begin{align*}
\label{eq: MMD}
    m_i &= \text{MMD}(X_f^i, X_p^i)  \\ 
        &= \frac{1}{\binom{w}{2}}\sum_{\substack{a,b=1 \\ a \neq b}}^w 2(K_{a,b}^i +  K_{-a,-b}^i) + \frac{1}{w^2} \sum_{a,b=1}^w 2K_{a,-b}^i. 
\end{align*}

Throughout this paper, MMD with a radial basis function kernel is used to detect change points unless otherwise specified. However, we again emphasize that any change point detection method  can be used as long as it performs well in identifying changes points.

The labeled data can be used to set the change point detection threshold $\tau$ and the window size $w$ to balance between false and missed change points. While we simply fix these parameters in advance using labeled data, these could also be considered as tuning parameters whose values can be set based on performance on a hold-out validation dataset.

\subsection{Pairwise constraints via change point detection}

Equipped with the detected change points, similar and dissimilar pairs of sub-sequences can be obtained in an unsupervised manner as shown in Figure \ref{Fig: pairs from cp}.
The idea is to form four consecutive non-overlapping sub-sequences. The first two sub-sequences  $( X_{p1} , X_{p2} )$ both occur before the change point. Since the change point detection algorithm did not determine that there was a change point in the combined segment of  $( X_{p1} , X_{p2} )$, we assume these two segments are generated by the same distribution and should be classified similarly. Similarly, the last two sub-sequences $( X_{f1} , X_{f2} )$ both occur after the change point and are also taken as a similar pair. In contrast, the segments on opposite sides of the change point have been identified as having different underlying distributions. In order to lead to a balanced distribution of similar/dissimilar pairs, we only use the constraints that  $(X_{f1} , X_{p2})$ and $(X_{f2} , X_{p1})$ should be classified differently.
Each of the subsequences above is chosen to be of a fixed length $s$ (determined by the spacing between change points).

\subsection{Clustered representations via pairwise constraints}

Using the approach described above, we can obtain similarity constraints from the unlabeled data. We can also obtain such constraints from labeled data via the assumption that sub-sequences corresponding to the same (different) class labels are similar (dissimilar) respectively. We can represent these as a set $\mathcal{P}_S$ consisting of sub-sequence pairs $(X_1,X_2)$ that are similar and a set $\mathcal{P}_D$ of dissimilar pairs. For compactness, we use the notation $P = (X_1,X_2)$ to refer to a sub-sequence pair belonging to $\mathcal{P}_S$ or $\mathcal{P}_D$.

These sub-sequences are then fed into a 1D temporal convolutional neural network \cite{bai2018empirical}, as illustrated in Figure \ref{Fig: NN diagram}. The neural network consists of 6 convolutional layers (or 3 temporal blocks as defined by \cite{bai2018empirical}) followed by 1 linear layer. We use a RELU activation function after every convolutional layer. We choose this architecture because the dilated filter structure leads to improved performance at classifying time series while being less computationally expensive than recurrent networks such as RNNs and LSTMs, although our framework could also easily accommodate either of these alternate network architectures.

Each instance $x_i$, in the input sub-sequence $X$, is passed through the neural network where the final linear layer transforms the output from the last convolutional layer into $\mathbb{R}^C$, where $C$ is the number of classes. A softmax function is then applied to obtain
the empirical distribution $f_{\theta}(x_i)$  for each instance $x_i$. For a length-$N$ sequence $X$, we define the mean empirical distribution as:
\[\widebar{f_{\theta}(X)} = \frac{1}{N}\sum_{i=1}^N f_{\theta}(x_i).
\]
We then compute the KL divergence between the mean empirical distributions for each sub-sequence within a pair $P = (X_1,X_2)$. Our loss function is constructed applying a hinge loss (with margin parameter $\rho$) to this KL divergence:
\[    h_{\theta}(P) = 
    \begin{cases}
      \text{KL}(\widebar{ f_{\theta}(X_1)},\widebar{ f_{\theta}(X_2) } ) & P \in \mathcal{X}_S, \\
      \rho  -   \text{KL}(\widebar{f_{\theta}(X_1)},\widebar{f_{\theta}(X_2)}) & P \in \mathcal{X}_D.
    \end{cases}
\]
The network is then trained according to the loss function:
\[
    \mathcal{L}_{\text{R}}(\theta) = \frac{1}{|\mathcal{P}_L|}\sum_{P \in \mathcal{P}_L} h_{\theta}(P) + \frac{\lambda_R}{|\mathcal{P}_U|}\sum_{P \in \mathcal{P}_U} h_{\theta}(P). 
\]
Here, $\mathcal{P}_L$ and $\mathcal{P}_U$ denote the sets of sub-sequence pairs in $\mathcal{P}_S \cup \mathcal{P}_D$ formed from the labeled and unlabeled data, respectively, and $\lambda_r$ is a tuning parameter which controls the influence of the unsupervised part of the loss function.

\subsection{Training a classifier}

Once trained, the network $f_{\theta}$ is fixed. The mean empirical distribution for an input sub-sequence $X$, $\widebar{f_{\theta}(X)}$, can then be used as a representation of $X$ that can serve as input to classifier network $f_{\psi}$. We use a 2-layer feedforward neural network followed by a softmax function to obtain a distribution over the  different classes.  
Labeled  as well as unlabeled sub-sequences (which correspond to the generated pairs from change points) are passed through this classification network. Since the learned representations encourage unlabeled data points to cluster around provided labeled data points,  known semi-supervised methods can be also used to incorporate unlabeled data while training $f_{\psi}$. We use entropy regularization \cite{grandvalet2005semi} to exploit the unlabeled data by encouraging the classifier boundary to pass through low density regions. 

The training data is comprised of two sets: $\mathcal{X}_L$ and $\mathcal{X}_U.$  Each element of $\mathcal{X}_L$ consists of a pair $(X,Y)$, where $X$ denotes a sequence $x_1, \ldots, x_N$ of vectors in $\mathbb{R}^D$ and $Y$ denotes a one-hot encoding of the class label for $X$ 
(and is hence in $\mathbb{R}^C$ where $C$ is the number of classes). Each element of $\mathcal{X}_U$ consists of a sub-sequence $X$ identified by the change point detection step (i.e., the individual sub-sequences in the set $\mathcal{P}_U$). The loss function that we use to train $f_{\psi}$ is given by:

\[
   \mathcal{L}_{\text{C}}(\psi) =  \frac{1}{|\mathcal{X}_L|} \sum_{ (X,Y) \in \mathcal{X}_L} \mathcal{L}_{\text{CE}}(X,Y)  + \frac{\lambda_C}{|\mathcal{X}_U|}\sum_{X \in \mathcal{X}_U} \mathcal{L}_{\text{NE}}(X).
\]
Here, $\lambda_C$ is a tuning parameter, $\mathcal{L}_{\text{CE}}$ is the cross entropy loss, and $\mathcal{L}_{\text{NE}}$ is the negative entropy loss:
\begin{align*}
\mathcal{L}_{\text{CE}}(X,Y) & = -\sum_{c=1}^C Y_c \log f_{\psi}\left(\widebar{f_\theta(X)}\right)_c  \\
     \mathcal{L}_{\text{NE}}(X) & = -\sum_{c=1}^C f_{\psi}\left(\widebar{f_{\theta}\left(X\right)} \right)_c \log f_{\psi}\left(\widebar{f_\theta(X)}\right)_c.
\end{align*}

Above, $f_{\psi}$ represents the output of the feedforward classification network which ends with a softmax distribution over $C$ classes. The input to $f_{\psi}$ is the mean empirical representations learned by network $f_{\theta}$ for input sequence $X$. The negative entropy loss   encourages the network $f_{\psi}$ to produce low entropy empirical class distributions  for unlabeled data.  This encourages unlabeled data to be mapped to a distribution that concentrates on a single class, pushing the classifier boundary to $f_{\psi}$ towards low-density regions.

A summary of our overall approach to semi-supervised learning via change point detection is given in Algorithm~\ref{Alg: 1}.


\begin{algorithm}
\begin{algorithmic}
\caption{SSL via change point detection}
\label{Alg: 1}
\State \textbf{Inputs:} Unlabeled sequence $X$, labeled sequences $\{X_l, Y_l$\}, CP detection parameters $\tau, w$,
\vspace{1mm}
\State \textbf{Output:} Trained networks: $f_{\theta}, f_{\psi}$
\vspace{1mm}
\State \textbf{Init:} Add similar/dissimilar pairs from $\{X_l\}$ to $\mathcal{P}_S, \mathcal{P}_D$ 
\vspace{1mm}
    \For{$i = 1$ to $\text{length}(X)$}
    \State Form windows: $X_p^i$, $X_f^i$   
    \State $m_i = \text{MMD}(X_p^i, X_f^i)$   
        \If{$m_i > \tau$}   
        \State Form two segments before CP: $X_{p1}^i,X_{p2}^i$  
        \State Form two segments after CP: $X_{f1}^i,X_{f2}^i$
        \State Add pairs $(X_{p1}^i,X_{p2}^i)$ and $(X_{f1}^i,X_{f2}^i)$ to $\mathcal{P}_S$ 
        \State Add pairs ($X_{f1}^i,X_{p2}^i$) and  $(X_{f2}^i,X_{p1}^i)$ to $\mathcal{P}_D$
        \EndIf
    \EndFor
    
    \For{ $j = 1$ to num\_epochs}
      \State Train  network $f_{\theta}$ by optimizing loss $\mathcal{L}_{\text{R}}$ 
    \EndFor
    \For{ $j = 1$ to num\_epochs}
      \State  Train network $f_{\psi}$ by optimizing loss $\mathcal{L}_\text{C}$ 
    \EndFor
\end{algorithmic}
\end{algorithm}

\section{Experiments}

\subsection{Baselines}

All of the following baselines use the same representation network $f_\theta$ and classification network $f_{\psi}$ architectures.

\subsubsection{Supervised}

In the supervised setting, only the labeled sequence is passed through through both the representation $f_{\theta}$ and classifier networks  $f_{\psi}$. We train the two networks in an end-to-end manner by minimizing:
\[
     \mathcal{L}_{\text{S}}(\theta, \psi) = \frac{1}{|\mathcal{X}_L|} \sum_{ (X,Y) \in \mathcal{X}_L} \mathcal{L}_{\text{CE}}(X,Y). 
     \label{eq : Supervised}
\]

\subsubsection{Denoising autoencoder}

A denoising autoencoder \cite{dai2015semi} or its variants such as the ladder network (where the reconstruction error for intermediate layers is also minimized) \cite{zeng2017semi} are often employed for semi-supervised learning with sequential data. Since it has been previously shown that the performance gap between these approaches is marginal \cite{zeng2017semi} -- which we have observed as well -- we focus only on the autoencoder as a baseline. In this approach, for every $X \in \mathcal{X}_U$, we also consider a perturbed version  $\widehat{X}$ produced by adding noise to $X$. Both $X$ and $\widehat{X}$ are passed through an encoder network $f_{\theta} $ to obtain embeddings which are used by a decoder network $f_{\theta}'$ to reconstruct the unlabeled data. A reconstruction loss of the form $\mathcal{C}(X) = \| X - f_{\theta}'(f_{\theta}(\widehat{X})) \|^2$ is incorporated into the loss function to exploit the unlabeled data. The labeled data is first passed through the encoder network $f_{\theta}$ to obtain embeddings, which are then fed into a classifier network $f_{\psi}$. We train the two networks in an end-to-end manner by minimizing:
\[   \mathcal{L}_{\text{AE}}(\theta, \psi) =  \frac{1}{|\mathcal{X}_L|} \sum_{ (X,Y) \in \mathcal{X}_L} \mathcal{L}_{\text{CE}}(X,Y) +
\frac{\lambda_C}{|\mathcal{X}_U|}\sum_{X \in \mathcal{X}_U} \mathcal{C}(X).
\]

\begin{figure*}[t!]%
    \centering
    \begin{subfigure}[\textbf{Autoencoder} ]{
        \centering
        \label{fig: MG_ae_30_semi}%
        \includegraphics[width=0.23\textwidth]{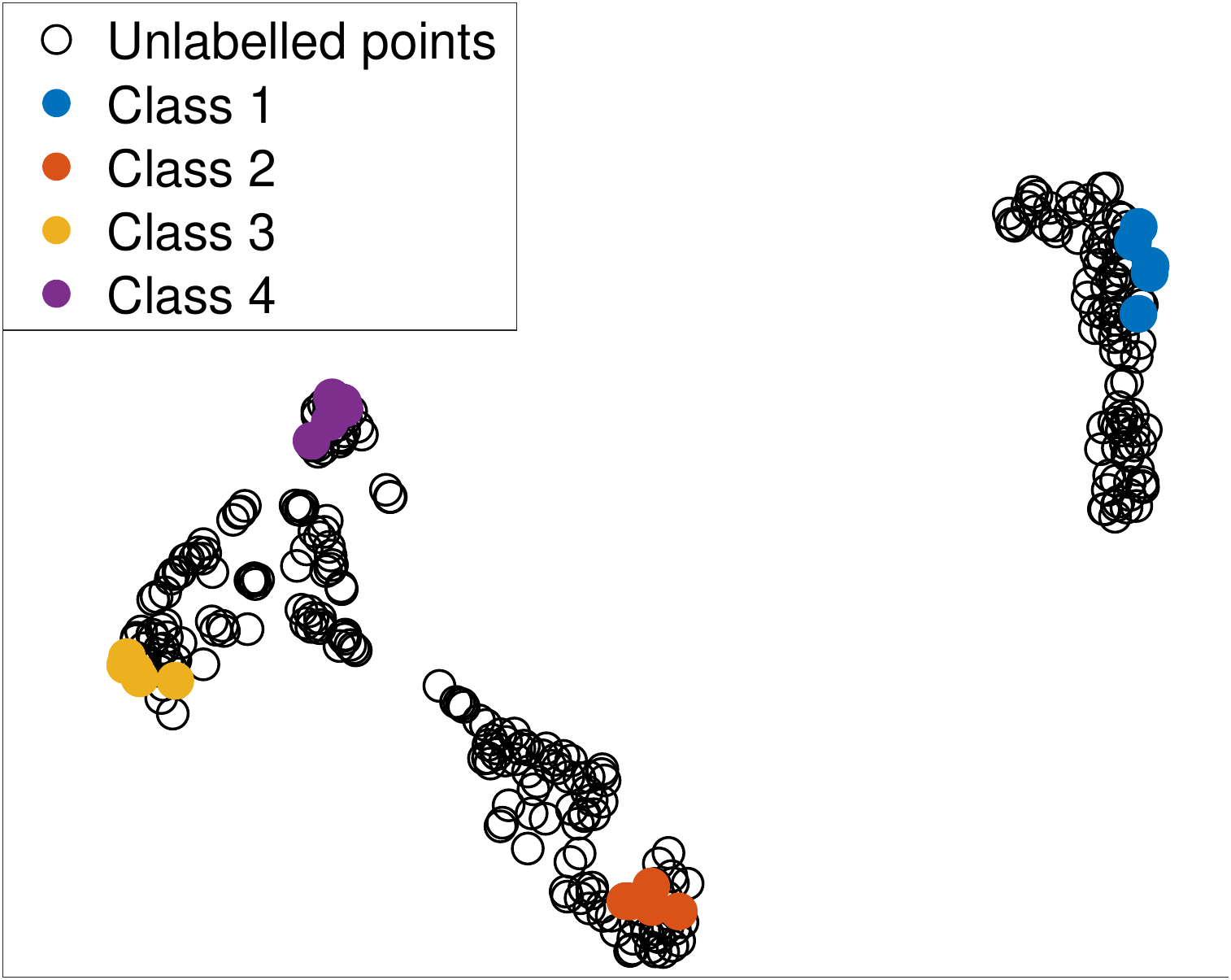}
    }\end{subfigure}
    \begin{subfigure}[\textbf{True labels: Autoencoder}]
     {   \centering
         \label{fig: MG AE all}%
         \includegraphics[width=0.23\textwidth]{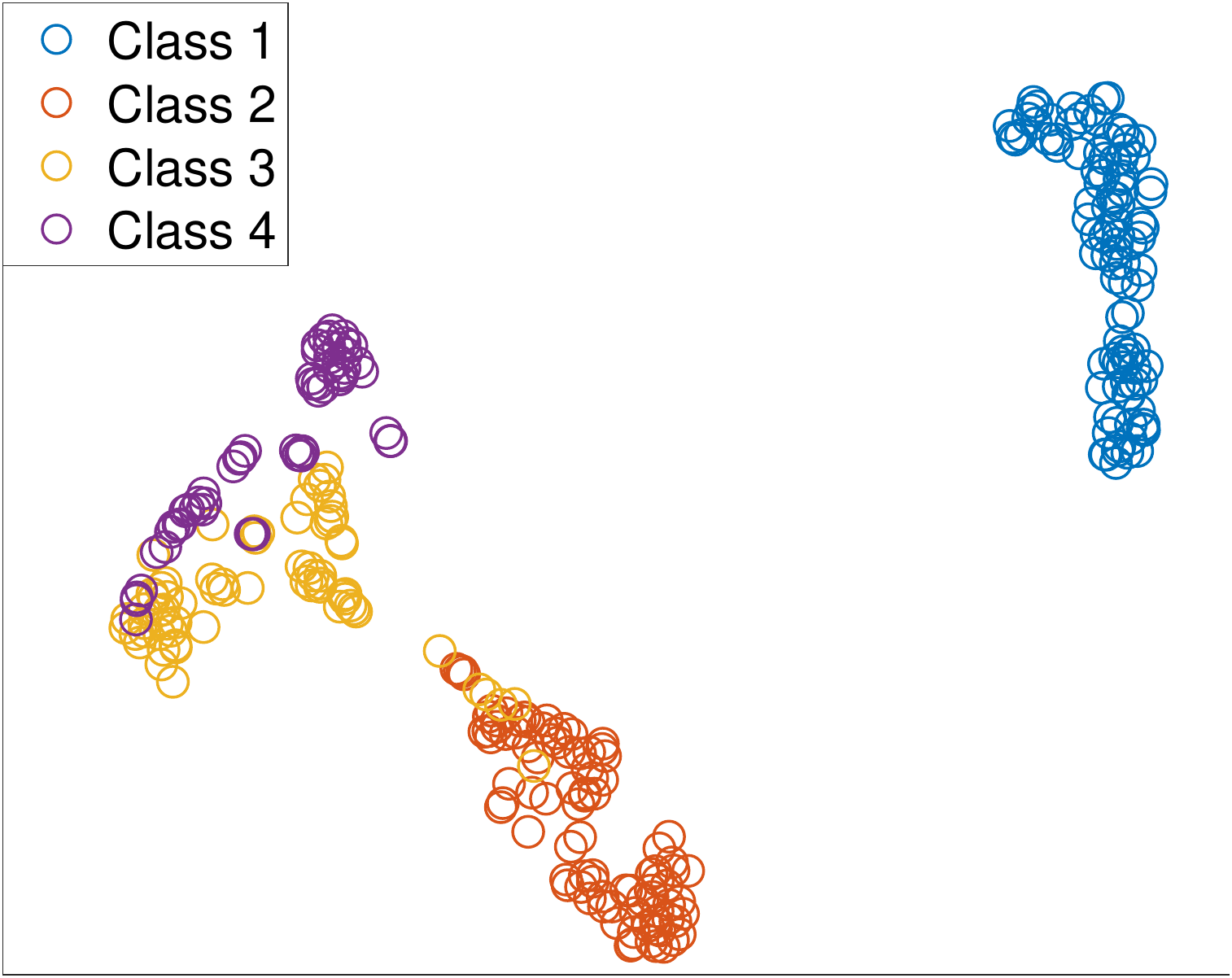}
    }\end{subfigure}
    \begin{subfigure}[\textbf{SSL-CP}]{ 
        \centering
        \label{fig: MG_cp_30_semi}%
        \includegraphics[width=0.23\textwidth]{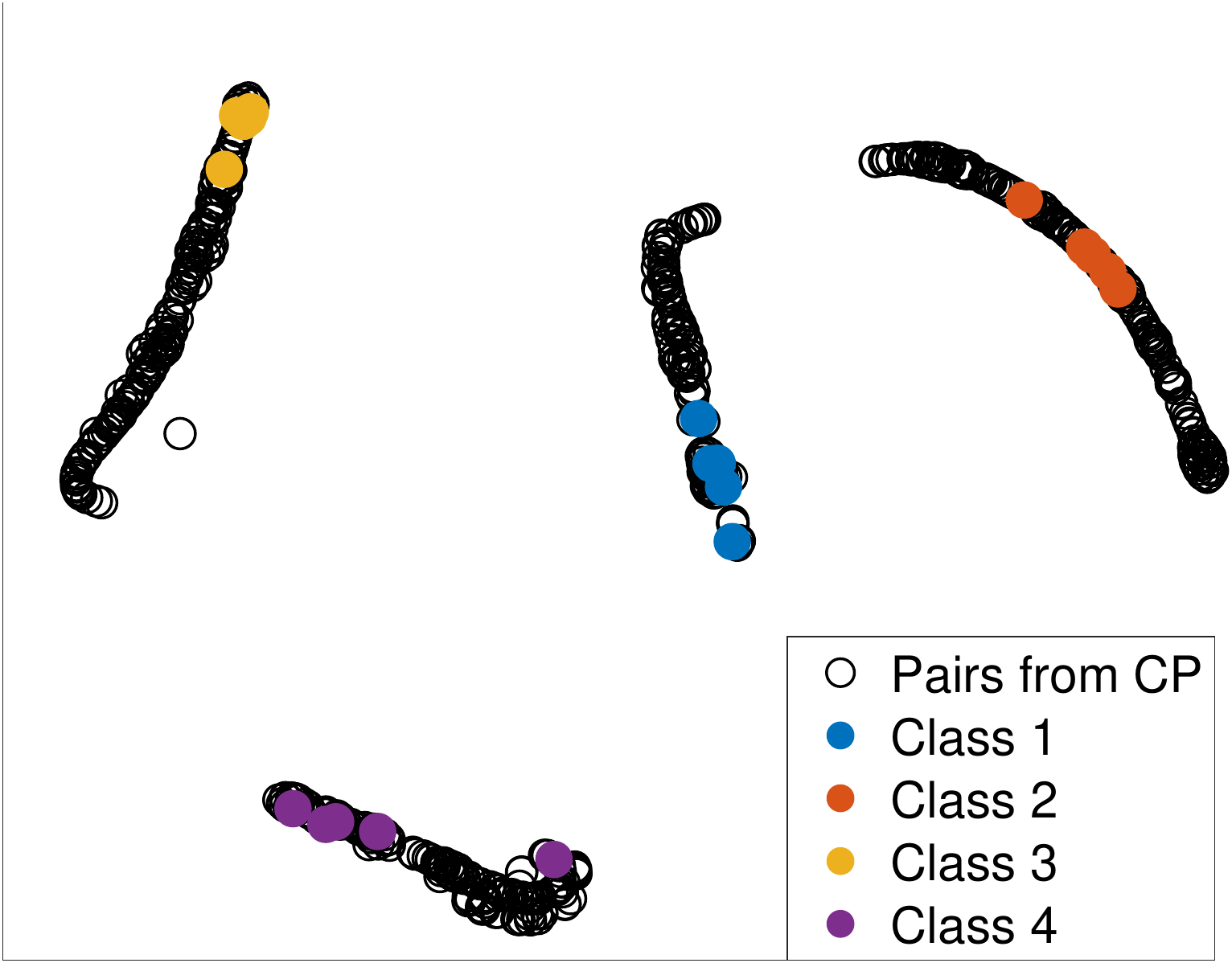}
    }\end{subfigure}
    \begin{subfigure}[\textbf{True labels: SSL-CP}]{ 
        \centering
        \label{fig: MG_cp_30_all}%
        \includegraphics[width=0.23\textwidth]{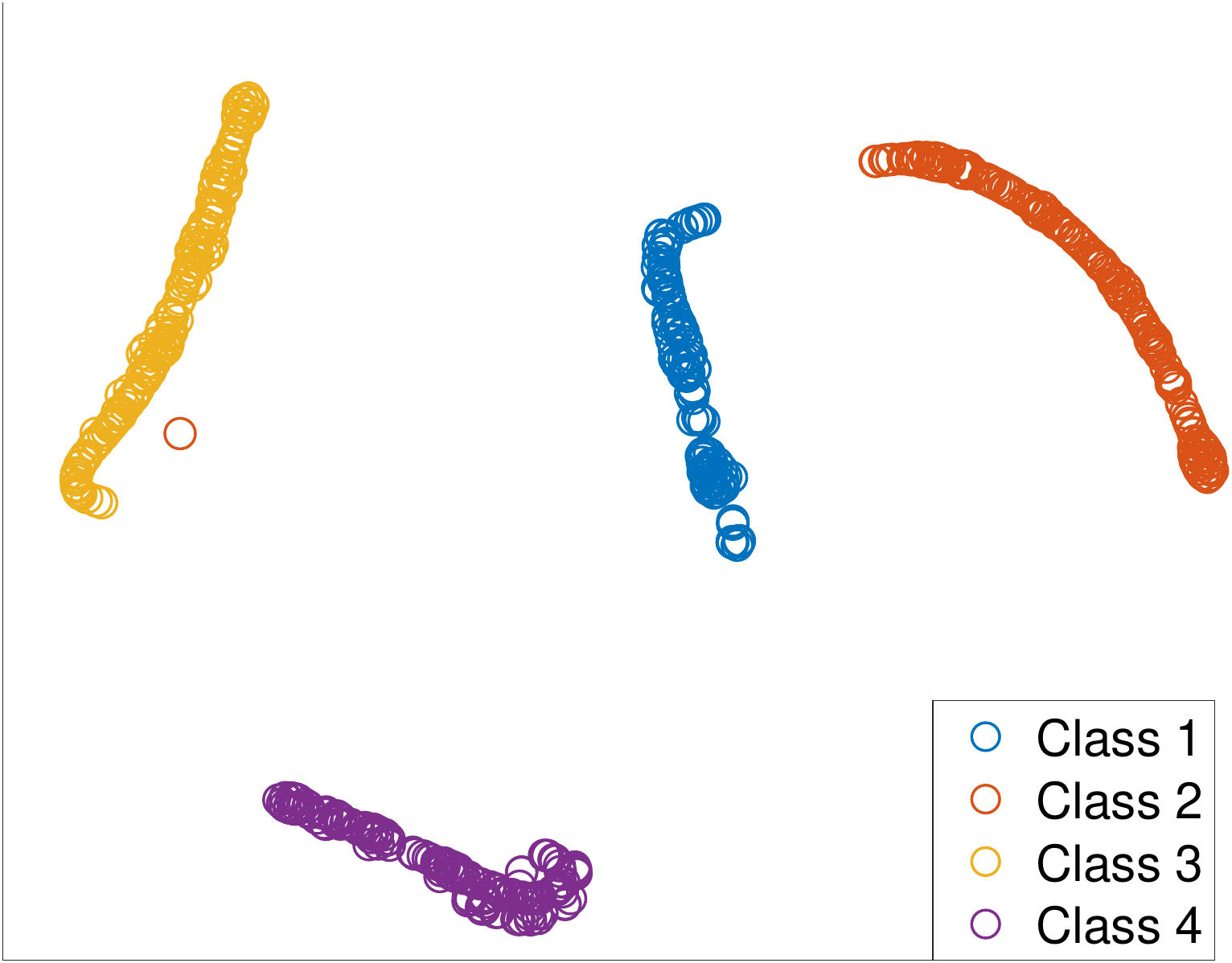}
     }\end{subfigure}
    \caption{T-SNE visualizations for the representations learned by the representation network ($f_{\theta}$) on the Mackay-Glass example when 5 labels are provided from each class. Figure \ref{fig: MG_ae_30_semi} shows representations learned by an autoencoder using both labeled and unlabeled data. It can be seen in \ref{fig: MG AE all} that different classes overlap in this  representation. Figure \ref{fig: MG_cp_30_semi} show the representations learned by SSL-CP, which are clustered and non-overlapping. This leads to improved classification when limited labels are provided. True labels for these representations are shown in Figure \ref{fig: MG_cp_30_all}.}
    \label{fig: Mackey-Glass representations}
\end{figure*}

\begin{table}[h]
  \caption{Classifier performance for mean, variance change}
  \centering
  \begin{tabular}{llll}
    \toprule
    
    Method  & 10 labels     & 30 labels   \\
    \midrule
     Supervised    &  0.90 $\pm 0.02$& 0.98  $\pm 0.01$  \\
    Autoencoder      &  0.87 $\pm 0.03 $ & 0.99 $\pm 0.01$     \\
    SSL-CP    &  0.99 $\pm$ 0.01 & 0.99 $\pm$ 0.01   \\
    SSL-CP (ER) & 0.99 $\pm$ 0.01 & 0.99 $\pm$ 0.01 \\
    \bottomrule
  \end{tabular}
  \label{Table: mean var}
\end{table}

\subsection{Synthetic experiments}

In all of the results below, we use the mean F1 score (unweighted) as an evaluation metric. In all synthetic simulations, we split the data in a 70/30 ratio where we use the larger split for training and the smaller split as a test dataset. We further split the training data in a ratio of 10/60/30. We use the smallest of these splits to obtain labeled data, the largest as unlabeled data for the semi-supervised setting, and the last split for validation. We use a small sub-sequence (comprising of 20 segments) in the unlabeled split to tune the parameters for change point detection. In our results, SSL-CP denotes our approach to semi-supervised learning via change points, but without the inclusion of the negative entropy term in the loss function.  SSL-CP (ER) denotes our approach when including this entropy regularization term.

\subsubsection{Changing mean and variance}

This example consists of data generated by a univariate normal distribution that switches its parameters  $(\mu,\sigma^2)$ every 500 samples.  We use 1500 such random switches to produce a sequence of data with five classes, correspond to the parameter settings $\{(2,0.1), (4,0.1), (4, 0.7), (10,0.1), (0,0.1)\}$. 
We use the symmetrical KL divergence from $\eqref{eq: Sym KL}$ to detect change points in the unlabeled data.
This is a simple change point detection problem where we detect all change points correctly. We use small sub-sequences of length 20 as labeled and unlabeled data. and we show the resulting performance in Table~\ref{Table: mean var}. This is a relatively simple sequence classification problem as it requires merely learning that the mean and variance determine class membership. 
Both the supervised and autoencoder baselines do reasonably well. However, classes 2 and 3 have the same mean but different variance, and both baselines struggle compared to SSL-CP in separating these classes when only 10 labels are provided.

\subsubsection{Mackay-Glass equation}

\begin{figure}[t]
  \centering
  \includegraphics[width=7cm]{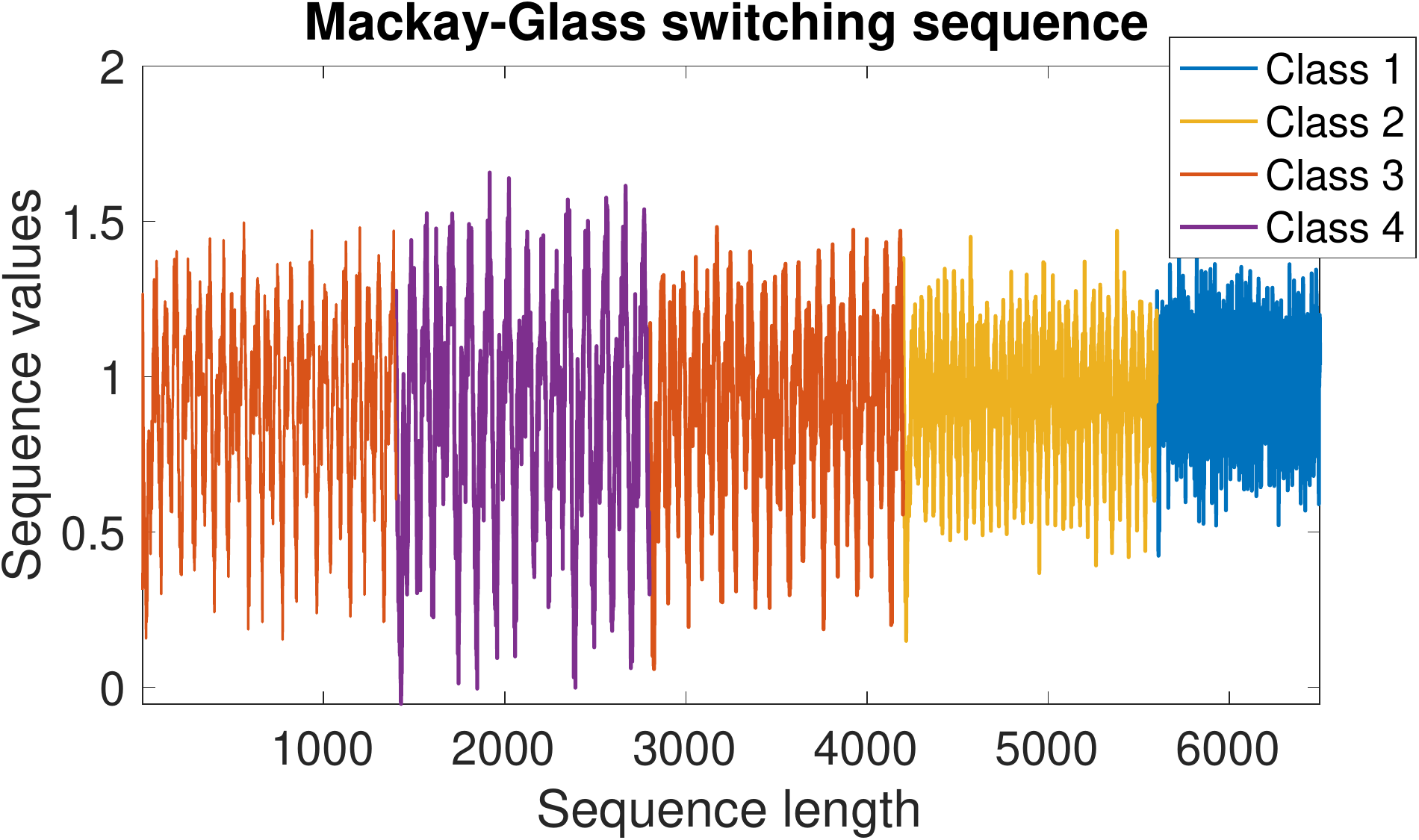}
  \caption{Example switching Mackay-Glass sequence.}
  \label{Fig: MG: sequence}
\end{figure}

The Mackay-Glass equation \cite{Glass:2010} is a non linear time delay differential equation defined as 
\[
  \frac{d(x(t))}{d(t)} =  -0.1 x(t)  +  \frac{\beta x(t) (t - \tau) }{1 + x (t - \tau)^{10} }.
\]
In a manner similar to \cite{kohlmorgen2002dynamic}, we generate a sequence by randomly switching  between parameters $( \beta, \tau) \in \{ (0.2, 8), (0.18, 16), (0.2, 22), (0.22,30)\}$ every 1400 samples. We define class membership according to the parameter settings of each segment. We generated 2000 such segments and added $\mathcal{N}(0,0.1)$ noise to the entire sequence. A small sub-sequence is shown in Figure~\ref{Fig: MG: sequence}.
We obtained pairs of sequences of size 100 using change points detected on the unlabeled dataset, where almost all true change points were detected correctly. There were about 4000 such pairs. We obtained 8100 non-overlapping windows of size 100 from the unlabeled-split for use by the autoencoder. Labeled data is also formed using non-overlapping windows of size 100 were used as labels. 
Table~\ref{Table: MG result} shows results for different numbers of provided labels.  We see that SSL-CP approach significantly outperforms the baselines. The representations learned by the autoencoder and SSL-CP are visualized in Figure \ref{fig: Mackey-Glass representations}, which illustrates that the autoencoder does not perform as well because it fails to learn representations that exhibit sufficient clustering. The influence of varying the number of provided pairs is shown in Table~\ref{Table: MG vary pairs}. We note that entropy regularization enhances the performance of SSL-CP when amount of unlabeled data is large.

\begin{table}[t]
  \caption{Mackay-Glass: Classifier performance for different number of labeled examples}
  \centering
  \begin{tabular}{lccc}
    \toprule
    Model  & 20 labels & 30 labels & 60 labels\\
    \midrule
    Supervised  &  0.55  $\pm 0.07$  & 0.86 $\pm$ 0.04 & 0.95  $\pm$ 0.02  \\
    Autoencoder & 0.73 $\pm$ 0.04 &0.90 $\pm$ 0.02 &0.98 $\pm$ 0.01\\
    SSL-CP & 0.96 $\pm$0.02 & 0.98 $\pm$ 0.01 & 0.99 $\pm$ 0.01\\
    SSL-CP (ER) &  0.99 $\pm$ 0.02& 0.99 $\pm$ 0.01 & 0.99 $\pm$ 0.01 \\
    \bottomrule
  \end{tabular}
  \label{Table: MG result}
\end{table}

\begin{table}[t]
  \caption{Mackay-Glass: Classifier performance for different amounts of unlabeled data}
  \centering
  \begin{tabular}{lccc}
    \toprule
    Model  & 600 Pairs & 1800 Pairs & 4000 Pairs\\
    \midrule
    SSL-CP &  0.87 $\pm$0.2   & 0.94 $\pm$0.1 & 0.96  $\pm$0.1\\
    SSL-CP (ER) &  0.87 $\pm$0.2  & 0.95 $\pm$0.1& 0.99 $\pm$0.2 \\
    \bottomrule
  \end{tabular}
  \label{Table: MG vary pairs}
\end{table}

\subsection{Real-world datasets}

\subsubsection{HCI: Gesture recognition}

The HCI gesture recognition dataset consists of a user performing 5 different gestures using the right arm \cite{forster2009unsupervised}. Data is obtained from 8 IMUs placed on the arm.  The gestures recorded included drawing triangle up, circle, infinity, square, and triangle down. We also consider the null case (where the user is not performing an activity) as a class. We use the free-hand subset from this dataset as it presents a relatively challenging  classification problem when compared with the more controlled subset. 
Rather than using consecutive non-overlapping windows (as the resulting sub-sequences are too small to contain a single class, since the duration of the null class can be very small), the sequential data is first divided into 100 segments using the labels.  30 segments are left as test data. 

This dataset presents a challenge to the SSL-CP approach in that most classes never appear adjacent to each other in the data set as they are always separated by a period in the null class. To obtain similarity constraints involving class pairs that do not include the null class, we generate a sequence by repeating a randomly sampled segment and concatenating it with another repeated randomly sampled segment. Change detection is then applied on this concatenated sequence to provide similar and dissimilar pairs. 600 of such similar dissimilar pairs were obtained.



When all labels within the dataset are provided, the mean F1 score for the supervised approach is 0.88. Such a score can actually sometimes be achieved by the supervised classifier even when only 1 label from each class is provided. However in this setting, the results can vary dramatically depending on exactly which instances are labeled. We obtained classification results across 30 trials, with a different random choice of which instance in each class were labeled. We show the average results in Table~\ref{Table: HCI}. In Table~\ref{Table: HCI diff seed} we show the percentage of trials in which each method performed best.

\begin{table}[t]
  \caption{HCI: Mean classifier performance when using one label per class}
  \centering
  \begin{tabular}{ccc}
    \toprule
    Supervised     & Autoencoder  & SSL-CP \\
    \midrule
     0.63 & 0.68  & 0.72   \\
    \bottomrule
  \end{tabular}
  \label{Table: HCI}
\end{table}

\begin{table}[t]
  \caption{HCI: Percentage of trials in which each method performs best when using one label per class}
  \centering
  \begin{tabular}{ccc}
    \toprule
    
    Supervised   & Autoencoder & SSL-CP \\
    \midrule
     11\% & 26\%  & 63\%   \\
    \bottomrule
  \end{tabular}
  \label{Table: HCI diff seed}
\end{table}
\subsubsection{WISDM: Activity recognition}

The WISDM activity recognition dataset \cite{forster2009unsupervised} consists of 36 users performing 6 activities which include running, walking, ascending stairs, descending stairs, sitting, and standing. Data is collected through an accelerometer  mounted on the participant's chest which provides 3 dimensional data sampled at 20Hz. For our experiments, we retained data from users 33, 34, and 35 as test set. We split the data from the rest of the users in a 70/30 ratio, using the large split for training and the small split for validation. We used a small sub-sequence (consisting of about 20 change points) to tune the change detection parameters. Once tuned, we obtained change points on the entire training set to obtain pairs of size 50. We obtained a total of about 4000 such pairs. We used about 7000 non-overlapping windows of size 50 as unlabeled data for the autoencoder. We used non-overlapping windows of size 50 as labeled data. In all experiments, we used a balanced number of labels from each class.

\begin{figure}[H]
  \centering
  \includegraphics[width=7cm]{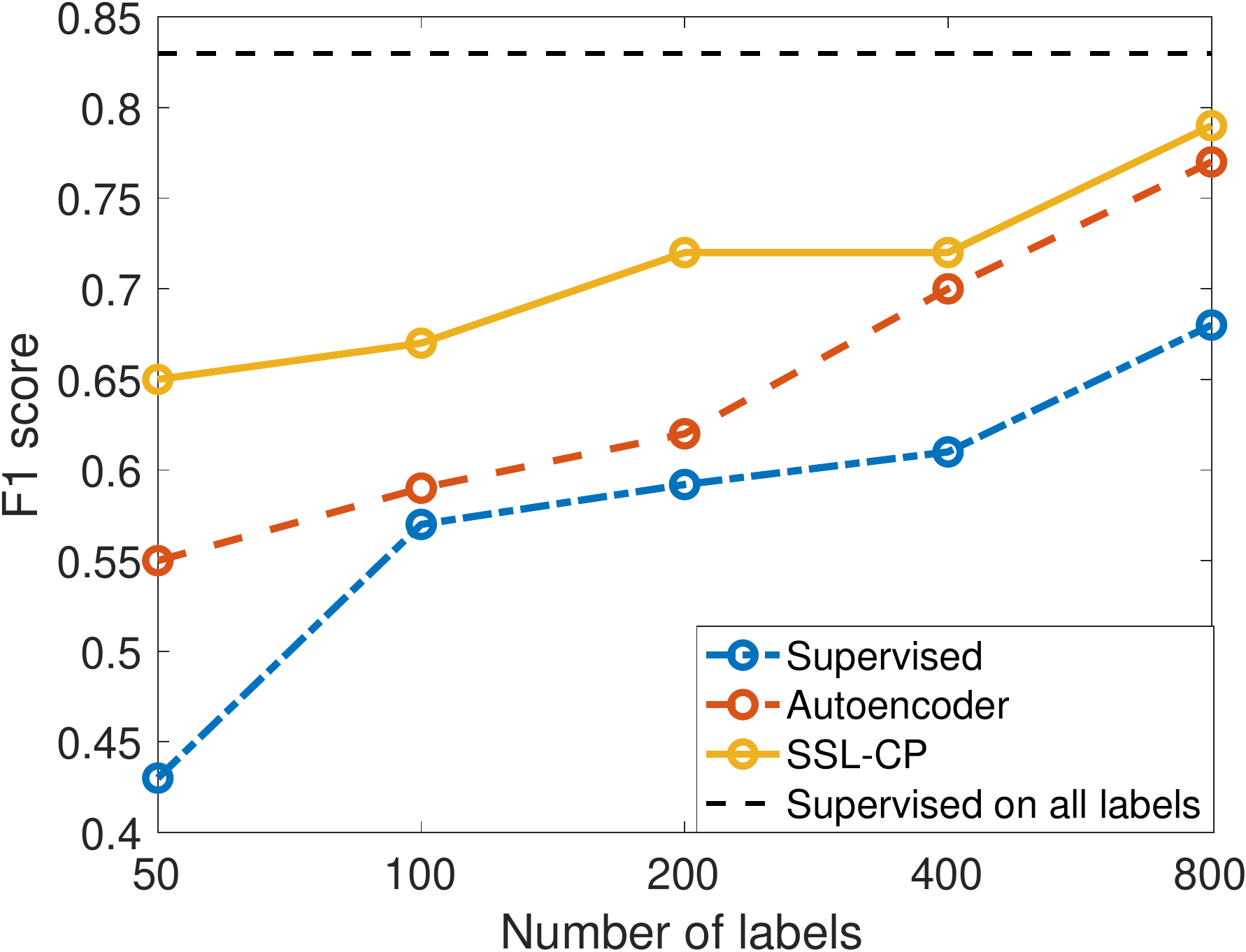}
  \caption{Performance on WISDM as the the number of provided labels increases. (Filtered users)}
  \label{Fig: WISDM label increase}
\end{figure}

Table \ref{Table: WISDM results} shows results when 48 labels (6 from each class). When pairs from all detected change points within the training set (4000 in number) are used, the performance of SSL-CP is slightly worse than that of the autoencoder. This is because many false change points are detected (up to about 40\% false change points) for a small number of users, leading to erroneous similarity constraints. After the removal of 10 such users, the number of falsely detected change points is reduced (to below 10\% across all users) and about 1600 pairs are obtained. The performance of SSL-CP for this case (filtered users) is notably better than the autoencoder. The performance further improves when all true change points are provided. In such a case, the number of unlabeled pairs are larger leading to improved performance of entropy regularization as well. Figure \ref{Fig: WISDM label increase} shows the relationship between classification performance and the number of labels available. In this experiment, only pairs derived from change points on the filtered users are used.

\begin{table}[H]
  \caption{WISDM: Classifier performance with 48 labels}
  \centering
  \begin{tabular}{lc}
    \toprule
    
   Method & F1 score \\
    \midrule
    Supervised  & 0.45  $\pm$ 0.04 \\
    Autoencoder &   0.54 $\pm$ 0.02 \\ \hline   \\[-3mm]  
     SSL-CP (All users) &  0.53 $\pm$ 0.03 \\
     SSL-CP (Filtered users) &  0.65 $\pm$ 0.02 \\
     SSL-CP (True CPs, all users) &  0.66 $\pm$ 0.01 \\
     SSL-CP-ER (Filtered users) &  0.65 $\pm$ 0.01 \\
     SSL-CP-ER  (True CPs, all users) &  0.69 $\pm$ 0.01 \\
    \bottomrule
  \end{tabular}
  \label{Table: WISDM results}
\end{table}

\newpage
\section{Discussion and Conclusion}

As highlighted by the performance on the WISDM dataset, the performance of our proposed method depends critically on the successful detection of change points. The detection of too many false change points can lead to corrupt similarity/dissimiarity constraints, that can potentially deteriorate performance. The other main limitation of the SSL-CP approach is that obtaining a rich set of similarity/dissimilarity constraints across all possible combinations of classes requires that these classes appear adjacent in the data. However, as we observed in the HCI dataset, the generation of additional sequences can provide a synthetic solution to this problem that is effective in practice.

Despite these limitations, SSL-CP consistently outperformed our baselines on  both synthetic and real-world datasets. This clearly shows the potential utility of incorporating information from change points in semi-supervised learning. Moreover, the results on the WISDM dataset clearly illustrate the potential improvement that could be realized by a more robust change point detection procedures.  Historically, change point detection has been mostly restricted to detecting anomalies or segmenting data. We hope that this work will encourage the community to recognize the utility of change point detection in semi-supervised learning and to devote more attention to developing improved non-parameteric change point detection procedures.

\bibliographystyle{plain}
\bibliography{Arxiv_SemiSupCP}

\newpage
\section{Appendix}

\subsection{Technical details on training neural networks}

\subsubsection{Preprocessing}

Inputs to all the networks are scaled to be between 0 and 1. 

\subsubsection{Network architecture and training details}
The neural network $f_{\theta}$  used to learn representations from change points consists of three temporal blocks, where each of the blocks consists of 100 channels. Here a temporal block is defined as in \cite{bai2018empirical} where each temporal block consists of two convolution layers with the same filter dilation. Each convolution layer is followed by weight normalization which is followed by a RELU activation and a dropout layer of 0.2.

For each successive temporal block, the filter was dilated by a factor of 2. 
The number of epochs needed to minimize training loss for $f_{\theta}$ was reduced by multiplying a constant (temperature \cite{zhang2018heatedup}) to the embedding provided to the final softmax layer. Details for these parameters can be found in Table \ref{Table: Hyperparams}. The first column refers to the different filters sizes (without dilation) used for each experiment. The second column lists the $\rho$ parameter for the hinge loss while the third parameter lists the temperature values used.

2000 epochs were provided to train $f_{\theta}$ through pairwise change points. The ADAM optimizer with a learning rate of 0.0001 was used for all experiments to train $f_{\theta}$.

\begin{table}[h]
\centering
\caption{Parameters for training $f_{\theta}$}
\begin{tabular}{| c | c  | c  | c | } 
      \hline
      Experiment & Filter size & Hinge param $\rho$ & Temp\\ 
       \hline
        Mean var &  5 & 4  & 5\\
    Mackay-Glass &  10 & 8 & 10\\
     HCI & 30 & 8 & 5\\
     WISDM & 10 & 4 & 10\\
    \hline
   \end{tabular}
   \label{Table: Hyperparams}
\end{table}
The feedforward fully connected network $f_{\psi}$ was trained using a learning rate of 0.001 through the ADAM optimizer and had two hidden layers of sizes 400 and 100 respectively. A RELU activation is used after each of these hidden layers.
400 epochs were provided to train the feed forward network.

For both the autoencoder and the supervised baselines, 1000 epochs were provided for training as the loss (for both validation and training, the loss became constant at the $700^{\text{th}}$ iteration and was constant until the $1000^{\text{th}}$ epoch.)

Each of the reported experiments was repeated 5 times, with mean and deviation (difference from the largest deviation from the mean) reported.
The seed for functions based on randomness was fixed to a value 5.

\subsection{Detecting change points}

Change points are also detected on sequences that are scaled between 0 and 1. This is not necessary but makes it convenient to  get scaled  similar/dissimilar pairs for the neural network $f_{\theta}$ directly from the sequence on which change points are detected.

\begin{figure}[h!]
  \centering
  \includegraphics[width=8cm]{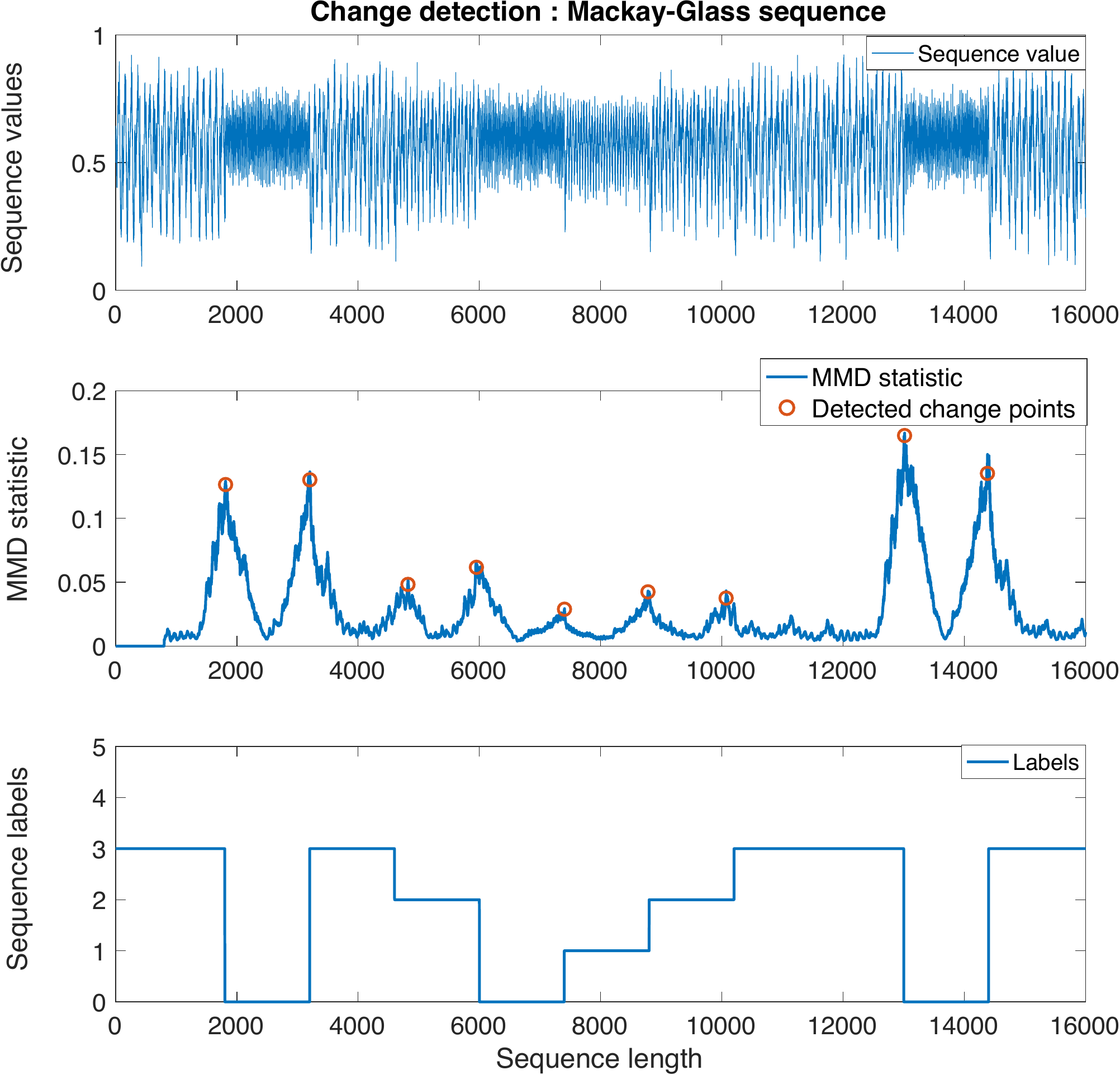}
  \caption{Detected change points on Mackay-Glass sequence}
  \label{Fig: Mackay CP}
\end{figure}

An example of detecting change points on the Mackay-Glass sequence can be seen in Figure \ref{Fig: Mackay CP}. This is a short sequence consisting of about 15 segments which can be used to set parameters needed for detecting change points. The first subplot shows the Mackay-Glass sequence. The second subplot shows the values of the MMD function as well as the detected changes while the third subplot shows the labels corresponding to different segments within the sequence. Note the mountain/hill  like features for the MMD statistic in the second subplot. These hills arise because the MMD function starts increasing when the future window $X_f$ starts overlapping with the segment belonging to the next class in the sequence. The peak value within this hill corresponds to the change point. The MMD function starts decreasing when the previous window $X_p$ starts overlapping with the sequence class corresponding to $X_f$ 

The peak function within the scipy \cite{2020SciPy-NMeth} python package can be applied on change statistics $m_i$ to obtain change points. The peak function is used with two options. One is the `peak height' which is equivalent to the change detection threshold $\tau$. The second argument is distance which specifies the minimum distance between detected change points. 

The parameters used for detecting change points are listed below. $\tau$ is the detection threshold, $w$ is the size of the windows for $X_f$ and $X_p$ and distance is the minimum distance between change points provided to the peak function.

\begin{table}[h]
\centering
\caption{Parameters for detecting change points}
\begin{tabular}{| c | c  | c  | c | } 
      \hline
      Experiment & $\tau$ & $w$  & distance\\ 
       \hline
        Mean var &  3 & 100  & 100\\
    Mackay-Glass &  0.025 & 800 & 800\\
     HCI & 0.18 & 600 & 600\\
     WISDM & 0.025 & 200 & 200\\
    \hline
   \end{tabular}
   \label{Table: Hyperparams}
\end{table}
\end{document}